\documentclass[letterpaper, 10 pt, conference]{ieeeconf}  

\IEEEoverridecommandlockouts                              
\usepackage{cite}
\usepackage{amsmath,amssymb,amsfonts}
\usepackage{graphicx}
\usepackage{textcomp}
\usepackage{xcolor}
\usepackage{algorithm}
\usepackage{algorithmic}
\usepackage{array}
\usepackage{tabularx}
\usepackage{booktabs}
\usepackage{multirow}
\usepackage{colortbl}
\usepackage{hhline}
\usepackage{color}

\title{\LARGE \bf
Optimizing 3D Gaussian Splatting for Sparse Viewpoint \\ Scene Reconstruction 
}

\author{Shen~Chen$^{1}$, Jiale~Zhou$^{1,*}$ and Lei~Li$^{2}$
\thanks{$^{1}$East China University of Science and Technology,
        {\tt\small zhou.jiale@ecust.edu.cn}
        }
\thanks{$^{2}$University of Washington,
        {\tt\small lilei@di.ku.dk}
        }
\thanks{$^{*}$Corresponding author.
}
}

\begin{document}

\maketitle
\thispagestyle{empty}
\pagestyle{empty}

\begin{abstract}
3D Gaussian Splatting (3DGS) has emerged as a promising approach for 3D scene representation, offering a reduction in computational overhead compared to Neural Radiance Fields (NeRF). However, 3DGS is susceptible to high-frequency artifacts and demonstrates suboptimal performance under sparse viewpoint conditions, thereby limiting its applicability in robotics and computer vision. 
To address these limitations, we introduce SVS-GS, a novel framework for Sparse Viewpoint Scene reconstruction that integrates a 3D Gaussian smoothing filter to suppress artifacts.
Furthermore, our approach incorporates a Depth Gradient Profile Prior (DGPP) loss with a dynamic depth mask to sharpen edges and 2D diffusion with Score Distillation Sampling (SDS) loss to enhance geometric consistency in novel view synthesis. Experimental evaluations on the MipNeRF-360 and SeaThru-NeRF datasets demonstrate that SVS-GS markedly improves 3D reconstruction from sparse viewpoints, offering a robust and efficient solution for scene understanding in robotics and computer vision applications.


\end{abstract}

\section{Introduction}
The use of RGB cameras in robotic vision systems for 3D scene reconstruction is essential for acquiring multiple viewpoints, a fundamental requirement for high-quality novel view synthesis (NVS). However, in practical scenarios, obtaining dense multi-view data is often impractical, especially in resource-constrained or complex environments. This limitation necessitates developing methods that can achieve effective scene reconstruction from sparse viewpoints. Traditional Neural Radiance Fields (NeRF) \cite{mildenhall2021nerf,barron2021mip,Barron_2023_ICCV} have shown strong performance in NVS, but their pixel-level ray rendering is computationally intensive and not well-suited for scenarios with sparse input data, requiring substantial resources and processing time.

In contrast, 3D Gaussian Splatting (3DGS) \cite{kerbl20233d} employs an explicit representation that significantly reduces both training and rendering times while maintaining high-quality outputs. This method initializes a set of 3D Gaussians from point clouds generated by Structure from Motion (SfM) \cite{schonberger2016structure} or via random initialization. It uses adaptive density control to clone and prune these Gaussians, enhancing scene detail representation. Leveraging the smooth, differentiable properties of Gaussian distributions, 3DGS enables rapid rasterization by projecting 3D Gaussians onto 2D image planes, supporting efficient rendering and interpolation \cite{kerbl20233d,tang2023dreamgaussian,yi2024gaussiandreamer}.

\begin{figure}[t]
   \centering
   \includegraphics[width=\linewidth]{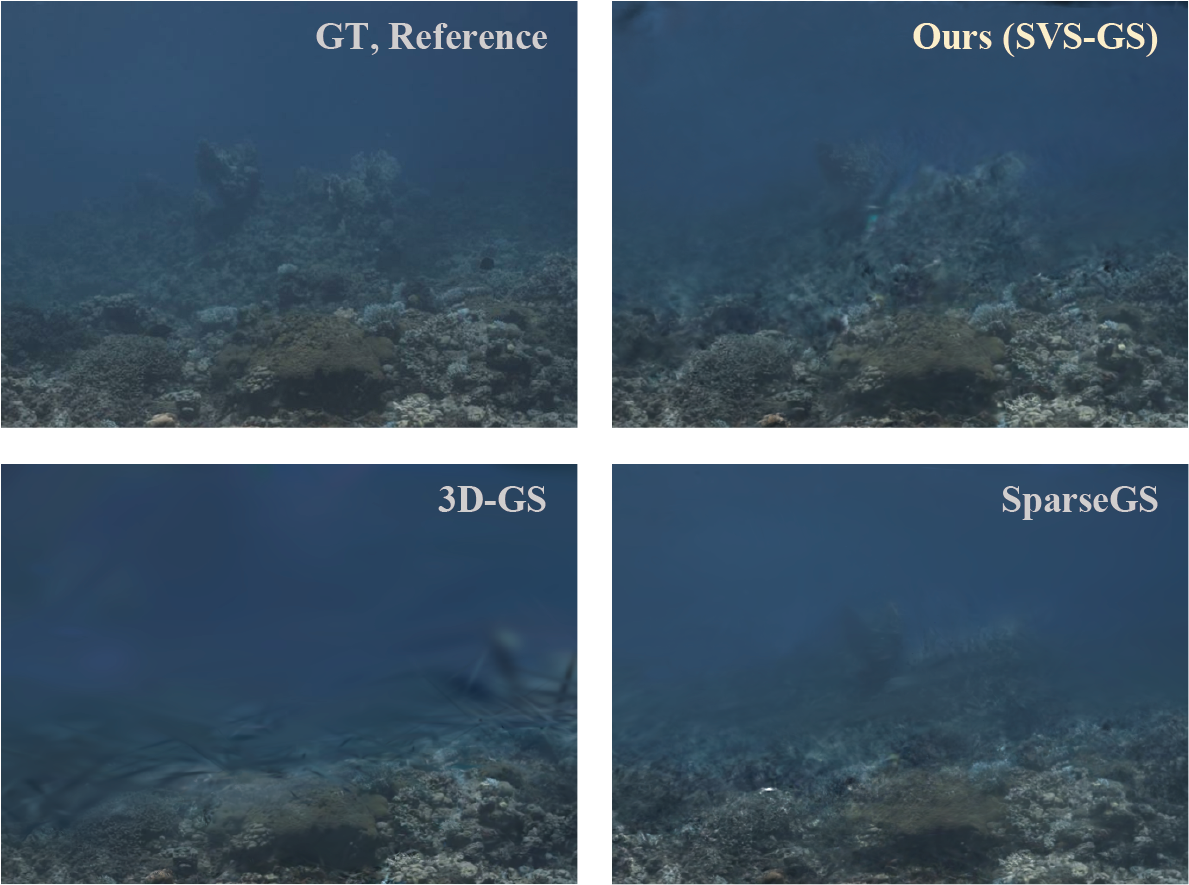}
   \caption{We propose a sparse Viewpoint scene Reconstruction framework.  Comparison of 3DGS \cite{kerbl20233d} and SparseGS \cite{xiong2023sparsegs} with our SVS-GS trained on 8 views shows that SVS-GS outperforms the other methods in synthesizing close-up scenes.}
   \label{fig:mov}
 \end{figure}

3D Gaussian distributions effectively capture details across multiple scales, and their projection onto a 2D plane simplifies the rasterization process. While this method is capable of efficiently representing complex, large-scale scenes or objects, the absence of size constraints for each 3D Gaussian primitive leads to a loss of detail when reconstructing fine objects, especially upon zooming in. This limitation is particularly evident when dealing with extremely thin lines, where it can result in inaccuracies that hinder the precise capture and reproduction of slender structures and small features, thereby compromising the overall visual realism and detail fidelity of the scene \cite{zhou2024feature,yu2024mip}. Moreover, in practical applications, 3D Gaussian Splatting (3DGS) requires densely sampled multi-view scenes to achieve optimal results \cite{zhang2024pixel,fan2024instantsplat,li2024dngaussian}. However, obtaining such extensive viewpoint data is often impractical in resource-constrained or complex environments. The unconstrained size of primitives in 3DGS and the reliance on dense multi-view image data present significant challenges for practical applications, such as autonomous vehicle navigation.


3DGS methods are heavily dependent on the density and quality of initial point clouds derived from dense multi-view inputs, which limits their effectiveness in sparse-viewpoint scenarios.  To address the inherent limitations of 3DGS, we propose a sparse-view 3DGS framework, termed SVS-GS. To impose size constraints on the 3D Gaussian primitives, we introduce a 3D smoothing filter \cite{yu2024mip}. This filter regulates the diffusion range of Gaussian primitives in both 3D space and their 2D projections, ensuring the preservation of more details during reconstruction, particularly for small and thin structures. In standard 3DGS, the initial 3D Gaussian primitives are derived from point cloud data generated by COLMAP \cite{schonberger2016structure,schoenberger2016mvs}. However, sparse views yield a limited number of initial points, resulting in low point cloud density, which adversely affects the distribution and quality of Gaussian primitives. To enhance the density of these initial 3D Gaussian primitives, we introduce a local adaptive density scaling module. This module dynamically increases the density of Gaussian primitives based on the sparse point clouds, producing a denser set of 3D Gaussian primitives.

For the optimization of the 3D Gaussian primitives, we employ score distillation sampling (SDS) loss \cite{poole2022dreamfusion} to integrate 3DGS with 2D diffusion, incorporating depth prior information to constrain the positions and sizes of the 3D Gaussian primitives. Additionally, we introduce a dynamic depth mask and Gradient Profile Prior (GPP) loss \cite{sun2010gradient} to enhance the sharpness of edges in the depth maps. SVS-GS effectively addresses gaps in the sparse point cloud data while simultaneously improving the uniformity and spatial coverage of the initial Gaussian primitives, thereby enhancing precision and detail fidelity in 3D scene reconstruction.

Our main contributions are as follows:
\begin{itemize}
    \item \textbf{Novel Sparse-View Framework}: SVS-GS reduces dependency on dense multi-view data by optimizing Gaussian primitive distributions, improving practicality and efficiency.
    \item \textbf{Adaptive Density Scaling}: A local adaptive density scaling module generates denser initial 3D Gaussian primitives, addressing the problem of sparse point clouds.
    \item \textbf{Enhanced Optimization Techniques}: Integration of SDS loss with 2D diffusion, dynamic depth masks, and depth priors ensures precise control over Gaussian primitives, improving detail reconstruction.
\end{itemize}


\begin{figure*}[ht]
   \centering
   \includegraphics[width=0.85\linewidth]{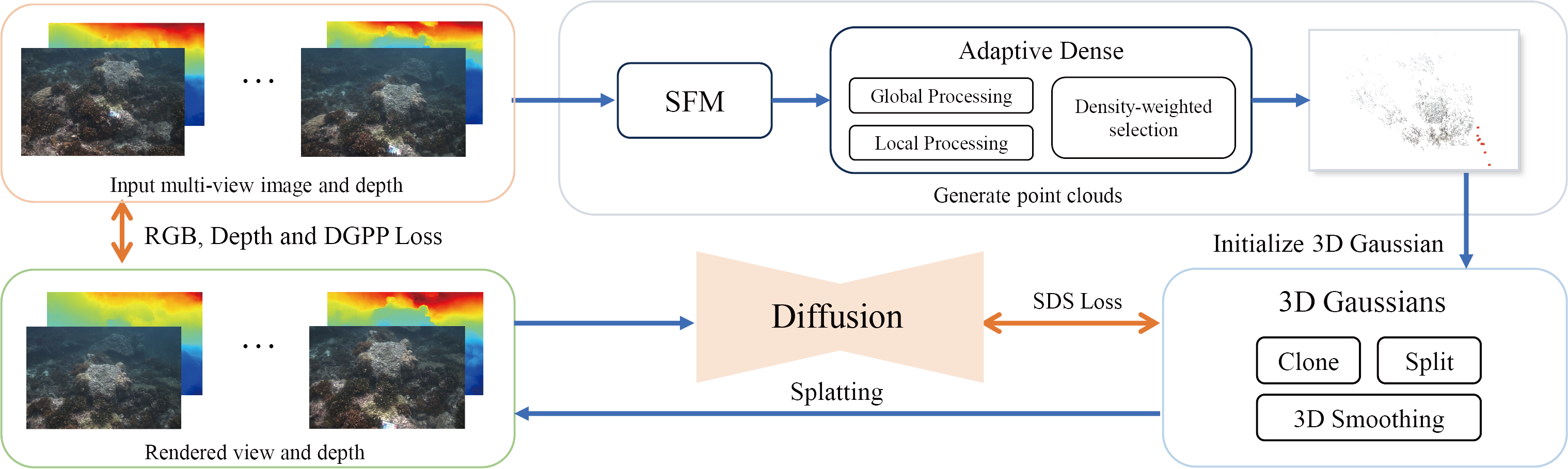}
   \caption{Overall framework. Starting with multi-view images and corresponding depth maps (obtained from Monocular Depth Estimation Models \cite{miangoleh2021boosting}), point clouds are generated by SfM and undergo adaptive density processing to optimize the density distribution of the point clouds. The point clouds are initialized as 3D Gaussian distributions and further refined through operations such as RGB, depth, and DGPP Loss. The SDS loss function is integrated to ensure geometric consistency and reduce noise.}
   \label{fig:Pipeline}
 \end{figure*}

\subsection{Novel View Synthesis}

Implicit representations for novel view synthesis (NVS), particularly Neural Radiance Field (NeRF)-based methods, have gained substantial attention in recent years \cite{mildenhall2021nerf,martin2021nerf,poole2022dreamfusion,tancik2022block,deng2022depth}. NeRF \cite{mildenhall2021nerf,martin2021nerf} utilizes a multi-layer perceptron (MLP) \cite{chen2019learning, mescheder2019occupancy} to predict radiance and density at 3D locations and viewing directions, leveraging classical volume rendering techniques \cite{kajiya1984ray} to generate high-quality novel views. Despite their strengths, these methods can produce artifacts when handling high-frequency details. To address this, Mip-NeRF \cite{barron2021mip} introduces multi-scale features and anti-aliased conical frustums to minimize blurring. While NeRF-based approaches are effective for objects and small-scale scenes, inaccuracies in camera parameters can accumulate errors in large-scale, unbounded environments, affecting reconstruction quality. Mip-NeRF 360 \cite{barron2022mip} alleviates these issues with non-linear scene parameterization and online distillation techniques to reduce artifacts in large-scale scenes.

In scenarios with sparse input views, NeRF models are prone to overfitting, which limits their ability to generalize to novel perspectives \cite{wang2023sparsenerf,yang2023freenerf}. Several methods have been proposed to enhance reconstruction accuracy in such settings. Depth-Supervised NeRF (DSNeRF) \cite{deng2022depth} combines color and depth supervision to produce more detailed scenes, while SPARF \cite{truong2023sparf} uses pixel matching and depth consistency loss to achieve high-precision 3D scene generation from sparse inputs.

\subsection{Primitive-Based Rendering}

Primitive-based rendering techniques, which rasterize geometric primitives onto a 2D plane, have gained widespread adoption due to their high efficiency \cite{grossman1998point,pfister2000surfels, zheng2023pointavatar}. Differentiable point-based rendering methods \cite{peng2021shape, yifan2019differentiable} are particularly effective for novel view synthesis (NVS) because they offer optimization-friendly representations of complex scene structures. Recently, the introduction of 3D Gaussian Splatting (3DGS) \cite{kerbl20233d} has renewed interest in explicit representation methods. Unlike implicit representations, explicit representations directly encode the geometry and lighting information of a scene, reducing computational complexity. However, 3DGS adapts Gaussian primitives to each training image independently, often neglecting the global structural coherence of the scene \cite{lu2024scaffold}. Additionally, the lack of size constraints during training can lead to artifacts in rendered novel views. To address these issues, Structured 3D Gaussians (Scaffold-GS) \cite{lu2024scaffold} introduces anchor points to guide the distribution of 3D Gaussian primitives, enhancing the structural integrity of the scene. Mip-Splatting \cite{yu2024mip} further improves 3DGS by incorporating a 3D smoothing filter and a 2D mipmap filter to constrain the size of Gaussian primitives, thereby capturing finer scene details.

Most 3DGS-based methods initialize using point clouds generated from Structure-from-Motion (SfM) techniques, such as COLMAP. These methods rely on dense input images to maintain sufficient point cloud density, which is crucial for high-quality scene reconstruction. When the input images are sparse, the resulting point clouds also become sparse, limiting the capacity of 3D Gaussian primitives to capture intricate geometric details during generation and optimization \cite{fan2024instantsplat,chen2024scalinggaussian}. This sparsity can cause the models to overfit to the limited training views, thereby hindering generalization to novel viewpoints and reducing the effectiveness of scene reconstruction. SparseGS \cite{xiong2023sparsegs} attempts to mitigate the dependency on dense input by incorporating 2D diffusion and depth information.


\section{Preliminaries}

3DGS employs anisotropic Gaussians to effectively capture the varying scales and orientations present within a scene. Each 3D Gaussian primitive, denoted as $\{ \mathcal{G}_n \mid n = 1, \ldots, N \}$, is characterized by several parameters: a center position $\mu_n \in \mathbb{R}^{3 \times 1}$, a covariance $\Sigma_n \in \mathbb{R}^{7}$, a color $c_n \in \mathbb{R}^{3}$, and an opacity $\alpha_n \in \mathbb{R}^1$. The Gaussian function is defined as:
\begin{equation}
 \mathcal{G}_n(x) = e^{-\frac{1}{2}(x-\mu_n)^T \Sigma^{-1}_n (x-\mu_n)},
\end{equation}
where $x$ denotes points queried around the center position $\mu_n$. The size and orientation of each 3D Gaussian primitive are determined by the semi-definite parameters $\Sigma_n = R_n S_n (R_n S_n)^T$, where $R_n \in \mathbb{R}^{4}$ represents a rotation matrix and $S_n \in \mathbb{R}^{3}$ is a scaling matrix.

To render images from different viewpoints, differential splatting is applied to project the 3D Gaussians onto camera planes. This process involves the viewing transformation $W_n$ and the Jacobian matrix $J_n$, resulting in a transformed covariance:
\begin{equation}
\Sigma'_n = J_n W_n \Sigma_n (J_n W_n)^T.
\end{equation}
For color construction, 3DGS utilizes spherical harmonics to model the color $c_n$ of each Gaussian, incorporating its opacity $\alpha_n$. When rendering from a novel viewpoint, the 3D Gaussians are projected onto 2D planes, and the resulting color $C_r(x)$ for a given ray $r$ is computed as:
\begin{equation}
C_r(x) = \sum_{i \in M} c_i \sigma_i \prod_{j=1}^{i-1} (1 - \sigma_j), \quad \sigma_i = \alpha_i \mathcal{G}^{2D}_i(x),
\end{equation}
where $c_i$ and $\alpha_i$ represent the color and opacity of the $i$-th Gaussian, respectively. Here, the ray $r$ originates from the camera center corresponding to the observation viewpoint. Finally, an adaptive density control mechanism is implemented to dynamically clone and prune the 3D Gaussians, maintaining a balance between computational efficiency and scene detail.

\section{Methods}

\subsection{Problem Formulation}

In the context of scene reconstruction, optimizing the initialized 3D Gaussian primitives necessitates a set of multi-view images \( I = \{I_1, I_2, \ldots, I_k\} \) and the corresponding point clouds \( P = \{ p_1, p_2, \ldots, p_M \} \). The multi-view images \( I \) are first utilized to generate an initial point cloud \( P \) through Structure-from-Motion (SfM) techniques. Subsequently, these images guide the optimization of 3D Gaussian Splatting (3DGS) by comparing them with the rendered images, thereby refining the 3D Gaussian primitives to improve scene representation. 

The quality of novel view synthesis (NVS) in 3DGS is heavily influenced by the density and distribution of point clouds \( P \) and the quality of the input multi-view images \( I \). When robotic vision systems rely exclusively on RGB cameras with limited data, the resulting sparse point clouds and input images can significantly impair the completeness and level of detail in the geometric representation, limiting the capacity of 3DGS to accurately capture scene complexity. This limitation becomes particularly critical in complex or unbounded environments, where inadequate data hampers the ability to represent intricate geometric structures and variations in lighting, thereby reducing the effectiveness of scene reconstruction.


\subsection{Initialize Adaptive Dense}
In 3D scene reconstruction, a combined strategy of global and local processing is employed to balance the accuracy of the overall structure with the refinement of local details. 
Global processing is responsible for capturing the broad geometric structure of the entire scene, while local processing focuses on enhancing the detail representation within specific regions.
\subsubsection{Global Processing}
The primary objective of global processing is to ensure the geometric consistency of the entire scene. Using the point clouds \( P_{\text{init}} = \{p_i \mid i = 1, \ldots, k\} \) generated by SfM, we first address the overall structure to obtain a comprehensive spatial framework and point cloud density distribution.
The global processing optimizes \( P_{\text{init}} \) to derive a global density function \( \rho(p) \):
\begin{equation}
    \rho_{\text{global}}(p) = \int_{P} \exp \left( - \frac{\|p - q\|^2}{2\sigma_p^2} \right) f(q) \, dq,
\end{equation}
where each point \( p_i \in P \) has coordinates \( (x_i, y_i, z_i) \), 
$q$ represents the potential nearest neighbors of the point $p$.
\( f(q) \) is the density function, representing the weight or density at point \( q \). 
This density function is utilized to assess the distribution of points across the point clouds, ensuring that the essential geometric structures are retained at the global level.

\subsubsection{Local Processing}
Following global processing, the point clouds are partitioned into several local regions \( N \), where each region undergoes more detailed optimization. The main goal of local processing is to enhance the representation of fine details. 
For a local region \( R_i \), the bounding box is defined as:
\begin{equation}
    p_{\text{min}_i} = \min(p_{R_i}), \quad p_{\text{max}_i} = \max(p_{R_i}),
\end{equation}
where \( p_{R_i} \) denotes the points within the region \( R_i \). The position of the newly generated points $p_{r} \in \left[ p_{\text{min}_i}, p_{\text{max}_i} \right]$ is determined by uniform sampling within this bounding box.
The local point cloud density function \( \rho_{local}(p) \) is further refined to capture intricate geometric details:
\begin{equation}
    \rho_{\text{local}}(p_r) = \int_{R_i} \exp \left( - \frac{\|p_r - q_r\|^2}{2\sigma_{p_r}^2} \right) f(q_r) \, dq_r,
\end{equation}
where \( R_i \) represents the integration domain, which encompasses the entire range of possible values for the local region around \( p_r \); 
$q_r$ represents the potential nearest neighbors of the point $p_r$.

\subsubsection{Density-weighted selection}
Upon completing the local and global density estimations, the point selection process strategically integrates these results, optimizing the balance between local precision and global coherence to enhance the overall quality of the reconstruction.

Initially, within each local region, a KD-tree \cite{bentley1975multidimensional} is constructed to identify the \( k \) nearest neighbors \( p_i \) for each point \( p \). 
The distances between \( p \) and these neighbors are calculated and then converted into local density values \( \rho_{\text{local}}(p) \) using a Gaussian function. Based on these density values, the probability of retaining each point \( \mathbb{P}_{\text{local}} \) is determined:
\begin{equation}
    \mathbb{P}_{\text{local}}(p_{r_j} \in p_{r}) \propto \rho_{\text{local}}(p_{r}).
\end{equation}
Simultaneously, a similar process is conducted at the global level. The global density \( \rho_{\text{global}}(p) \) is estimated by calculating the distances to the global nearest neighbors \( p_i \), and the corresponding global retention probability \( \mathbb{P}_{\text{global}} \) is computed:
\begin{equation}
    \mathbb{P}_{\text{global}}(p_{i} \in p) \propto \rho_{\text{global}}(p).
\end{equation}

The selected points from both the local \( P_{\text{local}} \) and global \( P_{\text{global}} \) density estimations are combined with the initial point cloud \( P_{\text{init}} \) using a union operation, resulting in the final point cloud \( P_{\text{final}} \):
\begin{equation}
    P_{\text{final}} = P_{\text{init}} \oplus P_{\text{local}} \oplus P_{\text{global}}.
\end{equation}
This approach ensures that both the global structural integrity and local detail accuracy are maintained, thereby improving the overall quality and precision.

\subsection{3D Smoothing}
The intrinsic and extrinsic parameters of the camera are not fixed, leading to varying degrees of artifacts when rendering novel views, especially upon magnification.
In the optimization process, the coordinates \( o_i = (x_{o_i},y_{o_i},z_{o_i}) \) of any arbitrary 3D Gaussian need to be transformed from the world coordinate system to each coordinate system of camera:
\begin{equation}
   e_i = o_i R_i + T_i = (x_{e_i},y_{e_i},z_{e_i}),
\end{equation}
where $R_i$ and $T_i$ represent the rotation matrix and  translation matrix for the i-th camera. The transformed point is then projected onto the image plane using the intrinsic matrix of the camera:
\begin{equation}
    x^s_i = \frac{x_{e_i}}{z_{e_i}} \cdot f_{i,x} + \frac{W_i}{2}, \quad y^s_i = \frac{y_{e_i}}{z_{e_i}} \cdot f_{i,y} + \frac{H_i}{2},
\end{equation}
where $f_i$ represents the focal length of the i-th camera; $H_i$ and $W_i$ represent the height and width of the image, respectively. 
The maximum Gaussian point frequency $\beta_k$ is obtained using the observed positions of the 3D Gaussians on the screen:
\begin{equation}
\begin{aligned}
\zeta_k = \sup \left( \frac{f_i}{z_{e_i}} \right), 
\end{aligned}
\end{equation}
where $x_i^s \in [-\alpha W_i, (1+\alpha) W_i]$ and $y_i^s \in [-\alpha H_i, (1+\alpha) H_i]$. The hyperparameter $\alpha$ is used to extend the boundary of the image plane, ensuring that points near the image edges are considered.

After 3D smoothing filtering, the 3D Gaussian is represented as follows:
\begin{equation}
    \mathcal{G}_k(x) = \sqrt{\frac{\Sigma_k}{\Sigma_{k_s}} } \cdot 
    e^{-\frac{1}{2}(\mathbf{x} - \mu_k)^T \Sigma_{k_s}^{-1} (\mathbf{x} - \mu_k)},
\end{equation}
where $\Sigma_{k_s} = \Sigma_k + \frac{s}{\zeta_k^2}\cdot \mathbf{I}$ represents the covariance matrix after filtering. 

\subsection{Depth SDS as Optimization Guidance}
Using the diffusion model to generate spatially aligned RGB images and depth maps, we can guide the 3DGS optimization process in both structure and texture.
The depth map for each view is computed by accumulating the depth values of \(\mathcal{N}\) ordered Gaussian primitives along the ray, using point-based \(\alpha\) blending:
\begin{equation}
    D_r(x) = \sum_{i \in \mathcal{N}} d_{\mu_i} \sigma_i \prod_{j=1}^{i-1} (1 - \sigma_j),
\end{equation}
where $d_{\mu_i}$ is the depth of the $i$-th Gaussian primitive center $\mu_i$ in the camera view. All depth maps from the training views are normalized for subsequent depth-based loss calculation.

We employ SDS \cite{poole2022dreamfusion,tang2023dreamgaussian} to guide the optimization of 3DGS through 2D diffusion \cite{rombach2022high}.
The rendered image \( \tilde{I} \) and depth map \( \tilde{D} \) from unseen viewpoints $v$ are jointly used to optimize 3DGS through SDS:
\begin{equation}
\begin{aligned}
\nabla_{\theta} \mathcal{L}_{\text{SDS}} &= \lambda_1 \cdot \mathbb{E}_{\epsilon_I, t} \left[ w_t \left( \epsilon_{\phi} (I_t; \tilde{I}^v,t) - \epsilon_I \right) \frac{\partial I_t}{\partial \theta} \right] \\
&+ \lambda_2 \cdot \mathbb{E}_{\epsilon_D, t} \left[ w_t \left( \epsilon_{\phi} (D_t; \tilde{D}^v, t) - \epsilon_D \right) \frac{\partial D_t}{\partial \theta} \right],
\end{aligned}
\end{equation}
where \( \lambda_1 \) and \( \lambda_2 \) are coefficients that balance the influence of image and depth; 
\( \epsilon_\phi(.) \) is the denoising function of 2D diffusion; 
\( \epsilon_I \), \( \epsilon_D \sim N(0, I) \) are independent Gaussian noises.
By integrating the 2D diffusion model, 3DGS can be optimized more effectively, enabling the generated images and depth maps from new viewpoints to more accurately reflect the geometric structure and textural details of the actual scene.


\subsection{Depth mask and Gradient Profile Prior}

Since noise and irrelevant details in the distant background can negatively impact the gradient calculation process, leading to blurred edges and loss of detail in the reconstruction, we introduce a dynamic depth mask to effectively suppress high-frequency noise and artifacts from distant objects, thereby improving the geometric accuracy and visual quality of the reconstruction. 
To accommodate scenes with varying depth distributions, $q_f$ for the far-distance threshold is calculated as follows:
\begin{equation}
    q_f = q_b + \left(\frac{\beta_D}{\beta_D + \alpha_D}\right) \times \Delta q,
\end{equation}
where \( \alpha_D \) and \( \beta_D \) represent the mean and standard deviation of the depth map \( D \), respectively. \( p_b \) is the base quantile, and \( \Delta p \) is the dynamic adjustment range. The generated mask \( M \) is defined as:
\begin{equation}
    M = \mathbb{1}_{D \leq T_f} = \mathbb{1}_{D \leq \text{Quantile}(D, q_f)},
\end{equation}
where \( \mathbb{1}_{(\cdot)} \) is an indicator function that assesses the visibility of depth map \( D \). The mask is determined by calculating the value \( T_f \) at the quantile \( p_f \) of the depth map \( D \).
The final masked depth map ($D_m = D \odot M$) is used for gradient operations.

The Depth Gradient Profile Prior (DGPP) is introduced to enhance the sharpness and accuracy of edges in the depth map, particularly focusing on refining the texture and geometric details.
The GPP loss is formulated to enforce the alignment of gradient profiles between the rendered depth map \( \hat{D}_m \) and the target depth map \( D_m \). 
When the pixel positions $b$ of \( \hat{D}_m \) and \( D_m \) correspond one-to-one, the DGPP loss function is defined as:
\begin{equation}
    \mathcal{L}_{\text{DGPP}} = \frac{1}{b_1 - b_0} \int_{b_0}^{b_1} \|\nabla \hat{D}_m(b) - \nabla {D}_m(b)\|_1 \, db,
\end{equation}
where \( \nabla \hat{D}_m \) and \( \nabla D_m \) represent the gradient fields of the rendered and target depth maps, respectively. 
The depth alignment ensures that the sharpness of edges is preserved and that the 3D reconstruction accurately reflects the underlying geometry.

\subsection{Loss Function}

To optimize the 3D Gaussian representation ($\{\theta_k=(\mu_k, \Sigma_k,\alpha_k, c_k) \}^K_k$), we designed a final optimization function that integrates various loss terms.
Our final loss function for optimizing 3D Gaussians is defined as: 
\begin{equation}
\begin{aligned}
    \mathcal{L}_{final} &=
    \underbrace{\mathcal{L}_{\text{RGB}}(\hat{I}(\theta),I) + \lambda_{\text{depth}} \mathcal{L}_{\text{depth}}(\hat{D}(\theta),D,M)}_{\text{loss of know view}}\\
    & + \underbrace{\lambda_{\text{SDS}} \mathcal{L}_{\text{SDS}}(\tilde{I}^v(\theta), \tilde{D}^v(\theta))}_{\text{loss of novel view}},
\end{aligned}
\end{equation}
where $\hat{I}$, $\tilde{I}^v$  represent the RGB images rendered by the 3D Gaussian primitives; $I$ represents the reference RGB image; $\hat{D}$, $\tilde{I}^v$ represent the depth maps rendered by the 3D Gaussian primitives; $D$ represents the reference depth map.

\section{Experiments}

\begin{figure}[t]
   \centering
   \includegraphics[width=0.95\linewidth]{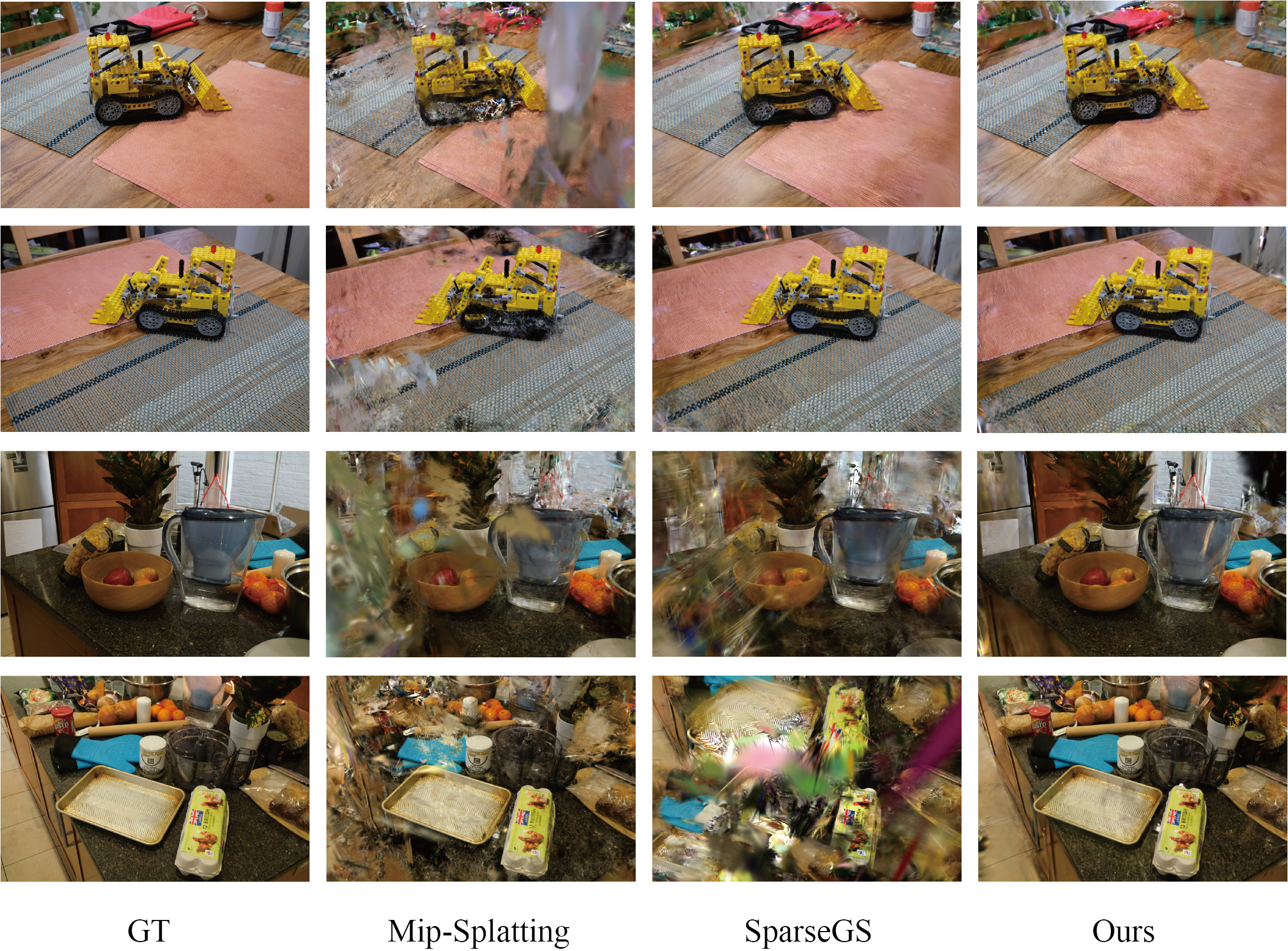}
   \caption{Qualitative results on the Mip-NeRF 360 dataset show that our approach is perceptually similar to the ground truth.}
   \label{fig:Qualitative1}
 \end{figure}

\begin{figure*}[t]
   \centering
   \includegraphics[width=0.85\linewidth]{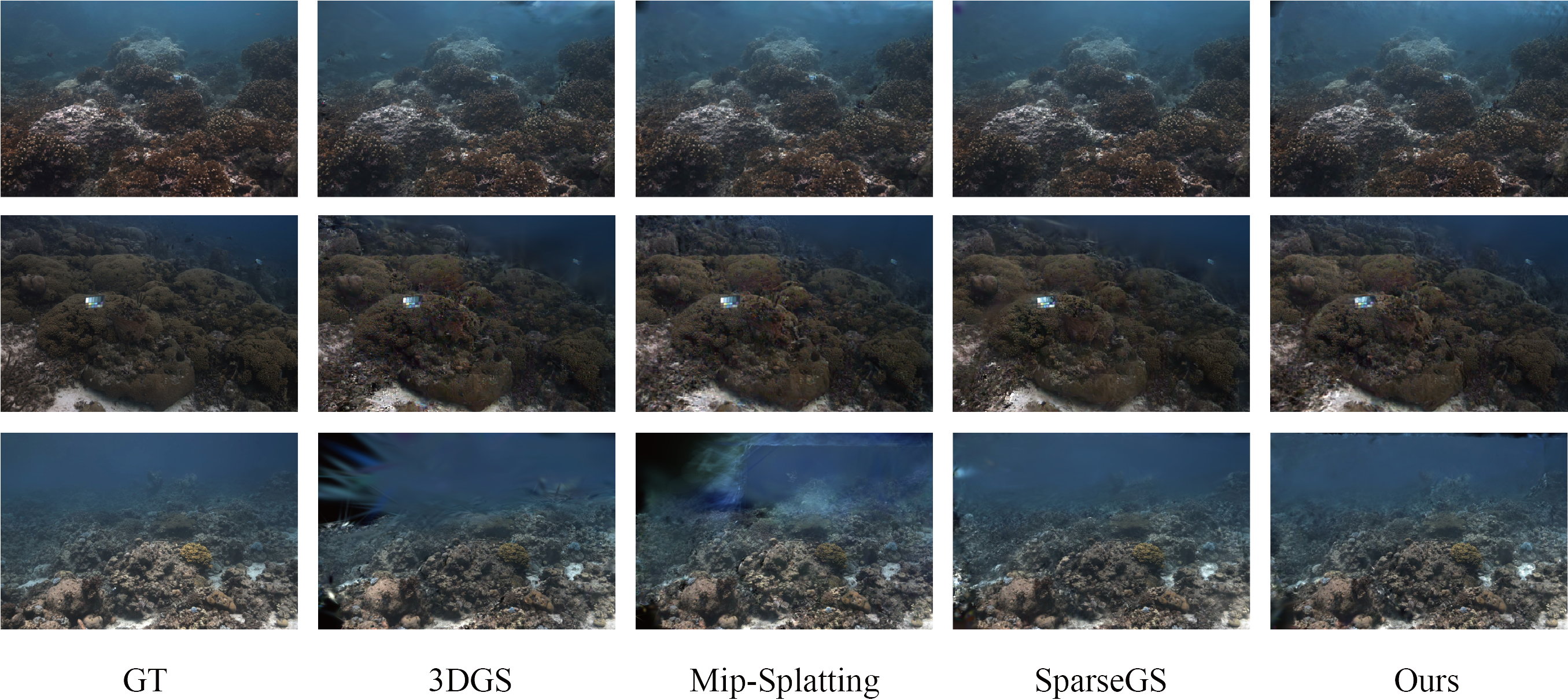}
   \caption{Qualitative results on the SeaThru-NeRF dataset show that our method can effectively shield the influence of distant scenery.}
   \label{fig:Qualitative2}
 \end{figure*}

\subsection{Datasets and Evaluation Metrics}
For unbounded scenes, we select six 360° coverage scenes from Mip-NeRF 360 \cite{barron2022mip} to evaluate our model. For underwater scenes, we used the SeaThru-NeRF dataset \cite{levy2023seathru} to evaluate the applicability of our framework to other complex scenes.
We employ tree metrics (PSNR, SSIM \cite{wang2004image}, and LPIPS \cite{zhang2018unreasonable}), to evaluate and compare our method against existing approaches.

\subsection{Implementation Details}
Our method is implemented using the PyTorch \cite{paszke2019pytorch} framework and the open-source 3DGS \cite{kerbl20233d} codebase. AdamW \cite{loshchilov2017decoupled} is employed as the optimizer. For all scenes, the models were trained for 30K iterations using the same loss function, Gaussian density control strategy, and hyperparameters to optimize the 3D Gaussian primitives. Both Gaussian training and rendering tests were performed on a $\text{NVIDIA}^{\text{TM}}$ RTX 4090 GPU.

\subsection{Qualitative and Quantitative Evaluation}


In the qualitative analysis on MipNeRF360, as shown in the Fig.\ref{fig:Qualitative1}, we compared the performance of different methods in reconstructing complex scenes. 
When reconstructing unbounded scenes, SVS-GS, with its integration of 3D Gaussian smoothing and depth priors, clearly outperforms traditional 3DGS and SparseGS methods by successfully capturing more intricate structures and lighting variations. These results further confirm the advantages and practical effectiveness of SVS-GS in sparse view scene reconstruction. Similarly, on the SeaThru-NeRF underwater dataset, SVS-GS again outperformed other methods, as shown in the Fig.\ref{fig:Qualitative2}. Particularly in handling the challenges of complex underwater lighting conditions and sparse viewpoints, SVS-GS demonstrated greater robustness and accuracy, successfully reducing visual distortions and preserving more scene details. These quantitative results underscore the broad applicability of SVS-GS across different scenarios and viewpoint conditions.
\begin{table}[t]
\centering
\caption{Quantitative Evaluation. Comparison of methods on Mip-NeRF360 and SeaThru-NeRF datasets. On the datasets with 8 input views, SVS-GS outperforms other methods.}
\resizebox{0.45\textwidth}{!}{%
\begin{tabular}{c|c|ccc|ccc}
\hline
\multirow{2}{*}{\textbf{Dataset}} & \multirow{2}{*}{\textbf{Method}} & \multicolumn{3}{c|}{\textbf{Mip-NeRF360}} & \multicolumn{3}{c}{\textbf{SeaThru-NeRF}} \\ \cline{3-8} 
& & \textbf{PSNR}$\uparrow$ & \textbf{SSIM}$\uparrow$ & \textbf{LPIPS}$\downarrow$ & \textbf{PSNR}$\uparrow$ & \textbf{SSIM}$\uparrow$ & \textbf{LPIPS}$\downarrow$ \\ \hline
\multirow{5}{*}{\textbf{Metrics}} 
& \textbf{Mip-NeRF360 \cite{barron2022mip}} & 11.28 & 0.193 & 0.612 & 22.89 & 0.830 & 0.245 \\ 
& \textbf{3DGS \cite{kerbl20233d}} & 10.45 & 0.163 & 0.640 & 22.24 & 0.785 & 0.242 \\ 
& \textbf{SparseGS \cite{xiong2023sparsegs}} & 12.33 & 0.225 & 0.593 & 22.99 & 0.769 & 0.234 \\ 
& \textbf{Mip-Splatting \cite{yu2024mip}} & 11.43 & 0.181 & 0.632 & 22.42 & 0.799 & 0.225 \\ 
& \textbf{Ours} & 12.80 & 0.238 & 0.573 & 23.06 & 0.791 & 0.214 \\ \hline
\end{tabular}%
}
\label{table:Quantitative}
\end{table}

In the quantitative analysis, we systematically evaluated the performance of SVS-GS against other methods on the MipNeRF360 and SeaThru-NeRF datasets, as shown in the Table.\ref{table:Quantitative}. Comparison of the PSNR, SSIM, and LPIPS metrics clearly demonstrates the significant advantages of SVS-GS in terms of reconstruction accuracy and image quality. On the MipNeRF360 dataset, SVS-GS achieved the highest scores in both PSNR and SSIM, indicating its superior ability to reconstruct geometric and textural details in sparse views, while also exhibiting the lowest perceptual error in the LPIPS, further validating its visual fidelity.

\subsection{Ablations and Analysis}
As shown in Table.\ref{table:Abl}, we conducted ablation studies to evaluate the impact of key components in our method. 
The dynamic depth mask plays a crucial role in effectively reducing noise and artifacts in distant areas, confirming its importance in filtering out irrelevant depth information.
DGPP sharpens edge contours, highlighting its importance in preserving details. 
Additionally, omitting the 3D Gaussian smoothing filter results in a noticeable increase in surface noise and artifacts, demonstrating its essential role in maintaining the smoothness and consistency of the reconstructed surfaces. The lack of SDS leads to geometric inconsistencies in the synthesized novel views, emphasizing the necessity of this component in ensuring geometric coherence and minimizing visual discrepancies.
Each component contributes to the effectiveness of achieving high-quality 3D scene reconstruction.

\begin{table}[t]
\centering
\caption{Ablation studies on underwater scenes. Comparisons on the SeaThru-NeRF dataset with 8 input views indicate that the model with all modules performs best, with each module contributing to the overall performance.}
\begin{tabularx}{0.45\textwidth}{l*{3}{>{\centering\arraybackslash}X}}
\toprule
 & \textbf{PSNR}$\uparrow$ & \textbf{SSIM}$\uparrow$ & \textbf{LPIPS}$\downarrow$\\
\midrule
w/o dense & 22.55 & 0.8155 & 0.2651 \\
w/o depth & 20.64 & 0.8115 & 0.2561\\
w/o 3D smoothing & 22.04 & 0.8130 & 0.2731\\
w/o SDS & 22.61 & 0.8189 & 0.2626\\
w/o DGGP & 22.72 & 0.8162 & 0.2663\\
All & 22.78 & 0.8234 & 0.2488\\
\bottomrule
\end{tabularx}

\label{table:Abl}
\end{table}

\section{Conclusion}
In this paper, we introduce SVS-GS, a novel framework for 3D scene reconstruction from sparse viewpoints, optimized for both robotic vision systems and broader computer vision tasks using only RGB cameras. Our method utilizes a dynamic depth mask to enhance geometric accuracy by selectively retaining critical depth information. Additionally, by incorporating depth priors, a 3D Gaussian smoothing filter, and Depth Gradient Profile Prior (DGPP) loss, our approach sharpens edges and preserves fine details in complex scenes. To ensure high-quality and consistent novel view synthesis, we integrate Score Distillation Sampling (SDS) loss, which reduces noise and maintains geometric coherence across different viewpoints. Experimental results demonstrate that SVS-GS outperforms existing methods in sparse viewpoint scenarios, achieving superior visual fidelity and geometric consistency. Furthermore, our framework shows robust performance across various challenging environments, making it an efficient and effective solution for 3D scene reconstruction in both robotics and computer vision applications.






\bibliographystyle{IEEEbib}
\bibliography{refs}

\end{document}